\documentclass{svproc}

\usepackage{url}
\usepackage{graphicx}
\usepackage{multicol}
\usepackage{multirow}
\usepackage{footmisc}
\usepackage{array}
\newcolumntype{L}{>{\centering\arraybackslash}m{3cm}}

\begin{document}
\mainmatter              
\title{Combining Multi-level Contexts of Superpixel using Convolutional Neural Networks to perform Natural Scene Labeling}
\titlerunning{Superpixel Level Natural Scene Labeling}  

\author{
Aritra Das\inst{1} \and 
Swarnendu Ghosh\inst{1}
Ritesh Sarkhel\inst{2} \and 
Sandipan Choudhuri\inst{3} \and 
Nibaran Das\inst{1} \and 
Mita Nasipuri\inst{1}}

\authorrunning{Aritra Das et al.} 


\tocauthor{Aritra Das, Swarnendu Ghosh, Ritesh Sarkhel, Sandipan Choudhuri, Nibaran Das , Mita Nasipuri}

\institute{
			Jadavpur University, Kolkata 700032, WB, India \and 
            Ohio State University, Columbus, OH 43210, USA \and 
            Arizona State University, Tempe AZ 85281, USA \\
            \email{\{dasaritra93|swarbir|sarkhelritesh|sandipanchoudhuri90\}@gmail.com,\\
            	{nibarandas|mitanasipuri\}@jadavpuruniversity.in}}
}

\maketitle              

\begin{abstract}
Modern deep learning algorithms have triggered various image segmentation approaches. However most of them deal with pixel based segmentation. However, superpixels provide a certain degree of contextual information while reducing computation cost. In our approach, we have performed superpixel level semantic segmentation considering 3 various levels as neighbours for semantic contexts. Furthermore, we have enlisted a number of ensemble approaches like max-voting and weighted-average. We have also used the Dempster-Shafer theory of uncertainty to analyze confusion among various classes. Our method has proved to be superior to a number of different modern approaches on the same dataset.
\end{abstract}

\section{Introduction}
Deep Learning has brought a new era in machine learning. Being able to learn more complex features from images, problems such as classification, localization, segmentation has seen remarkable progress especially for natural images. Previously most significant research in the domain of natural image processing was performed using some sort of pattern recognition over pixels ~\cite{oldsurvey1,oldsurvey2,oldsurvey3}. The problem that has been dealt in this paper is semantic image segmentation. Image segmentation goes beyond tasks like object recognition or localization. In this problem we are mainly interested in precise segments which semantically separates one object from another. While pixel level algorithms ~\cite{segreview1,segreview2,segreview3} provide very fine level segmentation, superpixels ~\cite{superpixelreview} provide much lesser computational complexity while not compromising performance. Superpixels refer to small patches of adjacent similar pixels grouped together. We have used these superpixels for our algorithms thus providing real-time performance. \textit{C}onvolutional \textit{n}eural \textit{n}etworks(CNNs) have showed tremendous performance in the field of natural image processing as well as segmentation. In our approach we have implemented multiple convolutional neural networks to obtain results. Any classification problem can be associated with uncertainty in the decision process. We have used some ensemble methods as well as Dempster-Shafer Theory to handle such uncertainty. The next section will give a brief review of some related works. Section 3 will explain the methodologies. In section 4 and 5 will cover the experimentations and discussions regarding obtained results.
\section{Related Works} 
Segmentation Algorithms also gained momentum with the onset of deep learning. In 2015, Ross Girshick in his paper F-RCNN ~\cite{girshick} outlined the quickest way of detecting multiple regions in an image. However, his proposed architecture does not segment the whole image but can find where the objects are in the image. In SegNet ~\cite{segnet}, the idea of convolution and the de-convolution have been used together to generated segmented regions. Farabet et al. \cite{farabet} showed how superpixel level classification may be performed by using CNNs. Though superpixels were only used to generate a scene parsing tree rather than considering them for the actual segmentation. Our approach however trains the CNN directly on the superpixel patches. While a variety of superpixels have been seen in the field of image segmentation ~\cite{superpixelreview}, our choice is the SLIC ~\cite{achanta2012slic}, for its speed and boundary adherence. Uncertainty is a common challenge in machine learning problems. In image segmentation we have seen the use of Dempster-Shafer Theory~\cite{shafer,shaferseg,shaferstudy} for elimination of such uncertainties as well. For our current the ICCV09 Dataset ~\cite{ICCV09} was used.
\section{Methodologies}
\begin{figure}[htbp]
\centering
\includegraphics[width=\textwidth]{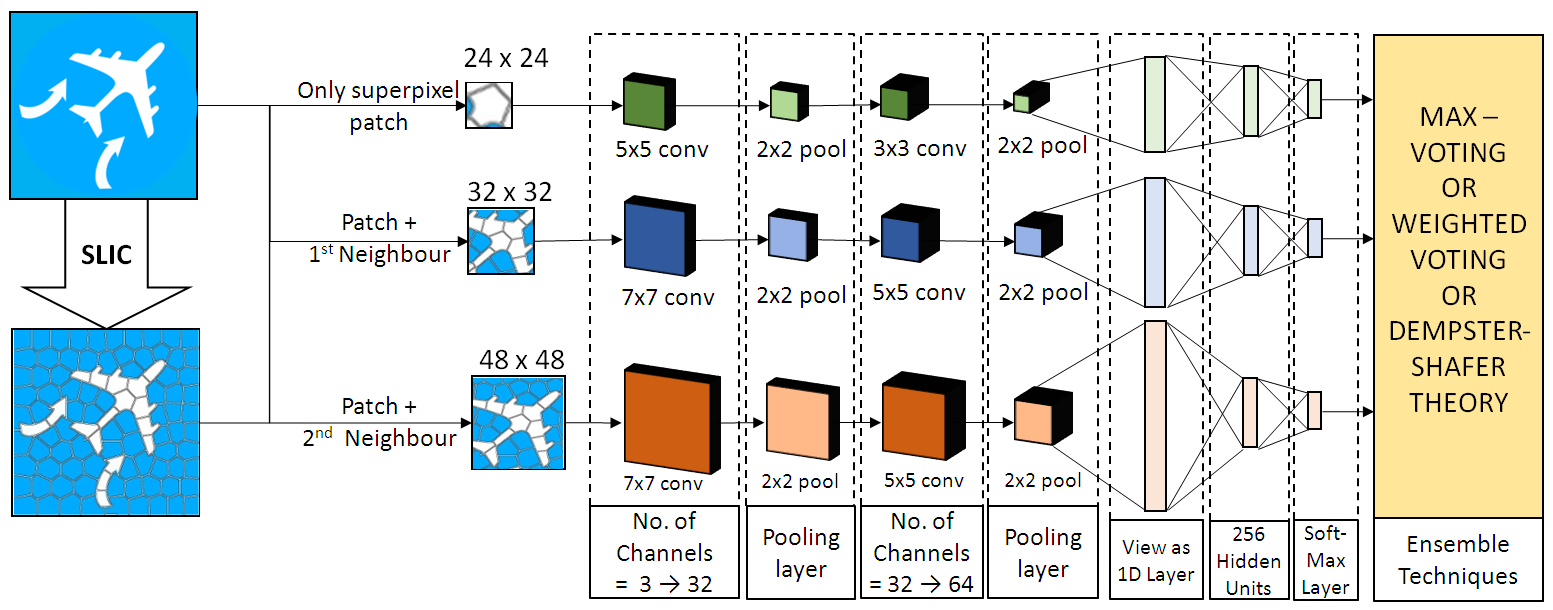}
\caption{Overall Flowchart}
\label{flowchart}
\end{figure}

First phase of our approach deals with training CNNs for classifying superpixels into 8 categories w.r.t the 8 semantically segmented classes of the ICCV 09 Dataset. Second phase  deals with the ensemble of three different variations of the CNN using various methods. The overall workflow is clearly demonstrated in fig \ref{flowchart}.
The following subsections will explain each module in details.
\subsection{Superpixel based Segmentation(Module 1)}
Pixel level classification is a tedious process primarily due to two factors. Firstly, even a small image contains quite a high number of pixels, and secondly, the information content of a pixel is very limited to consider classification into various segments. By using superpixels, we capture much more information than a single pixel and number of superpixels in an image is much lesser than the number of pixels. First  each image was divided into superpixels by using SLIC ~\cite{achanta2012slic}. To keep uniformity in the sizes of superpixels across images of various sizes of images, the \textit{m}inimum \textit{o}bject \textit{r}esolution $MOR$ was fixed. The \textit{n}umber \textit{o}f \textit{s}uperpixels $NOS$ of an image $I$ is given by,

\begin{equation}
  NOS(I) = \frac{Height(I) * Width(I)}{MOR}.
\end{equation}

\begin{figure}[htbp]
\centering
\includegraphics[width=\textwidth]{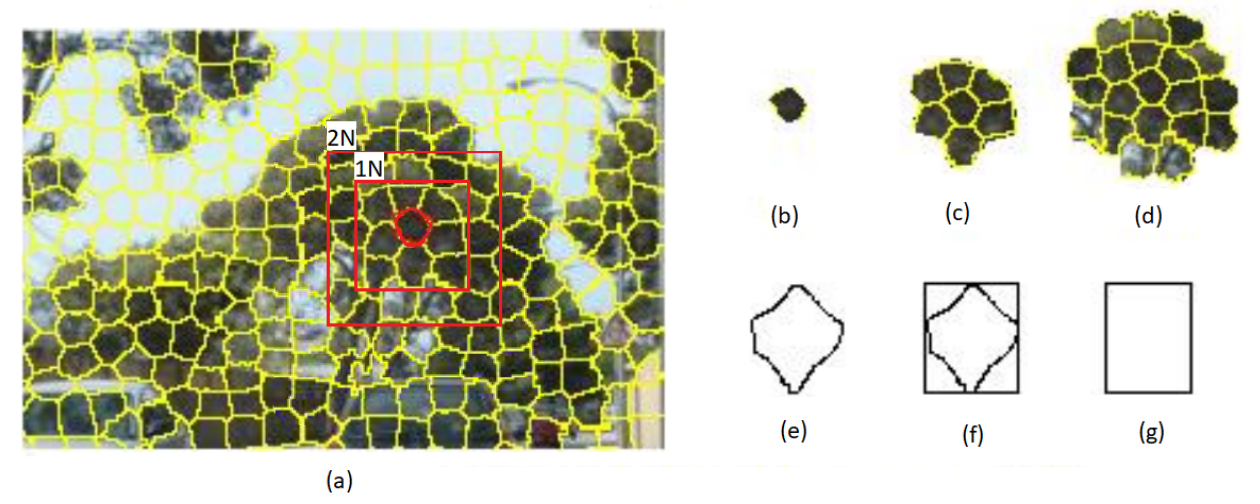}
\caption{Superpixel’s Neighbor. (a) Image with selected superpixel (in red), (b) Single Superpixel patch, noted as 0N, (c) Superpixel patch with 1st neighbor, noted as 1N (d) Superpixel patch with 2n d neighbor, noted as 2N, (e) single superpixel patch or patch with neighbors cropped out from image, (f) Minimal covering bounding box, (g) Regular size cropped out patch fed into the CNNs}
\label{superpixel}
\end{figure}

A patch of superpixel shows significant amount of texture information with respect to just pixels. However for semantic segmentation we also need to consider the context in which this superpixel occurs. So each superpixel was augmented with its neighbours to create a larger patch for the CNN to extract features from. For our experiments we have considered the first, second and third neighbour of each superpixel for its classification. Three different CNNs were trained for each of this neighbour category. It can be clearly seen in fig. \ref{superpixel}
Each CNN for classifying the superpixels consisted of two layers. First layer has $32, 5 \times 5$ convolution kernels followed by a standard $2 \times 2$ pooling. The second layer consists of $64, 3 \times 3$ convolution kernels followed by a standard $2 \times 2$ pooling. This is followed by a fully connected layer with 256 hidden units and a softmax output layer. 
\subsection{Ensemble Strategy}
Each of three CNN outputs a 8-dimensional softmax distribution. These are ensembled using three different methods, namely, max-voting, combination of mass function with the help of Dempster-Shafer theory of uncertainty and weighted sum techniques. 
\paragraph{Max-Voting:} This techniques takes the three predictions from three CNNs and chooses the winner on the basis of votes. In case of a tie the prediction with highest score was chosen. 
\paragraph{Weighted Average:} For weighted average, the output score is calculated by a weighted combination of all the softmax scores. The weight is determined by the training performance of each CNN. The final score $S_i$ for patch $i$ is given by,

\begin{equation}
  S_i = \frac{S_i^1 * r_1 + S_i^2 * r_2 + S_i^3 * r_3}{r_1+r_2+r_3}
\end{equation}

\paragraph{Dempster-Shafer Theory of Uncertainty:} There are certain number of superpixels for which the network gives poor predictions. The uncertainty rises in training because of lots of reasons like skewed datasets, similar superpixels for different classes, wrong ground truth level annotations. To deal with such uncertainties Dempster-Shafer [50] theory of evidence is taken into account. Unlike normal classification which uses a probability distributions across the number of classes, Dempster-Shafer theory of uncertainty deal with the masses and beliefs which are distributions across the all the possible combination of the classes. Hence forth mass value of a certain combination is defined by $m_j, where j \in 2^C and C = no. of classes$. So we designed an approach to simulate mass distribution by using the confusion matrix obtained during training. In theory of evidence $mass(A) <= Prob(A) <= Bel(A)$. The power set $\Phi(C)$ is written as 
\begin{equation}
\Phi(C) = \{m_1,m_2,...,m_C,m_{11}, m_{12},...,m_{1C},...,m_{12...C}\}
\end{equation}
So this difference in the between the mass value $m_i$ and probability $p_i \forall i \in C$  is defined in terms of the confusion (misclassification) related to that class. 

\begin{equation}
m_i = p_i - p_i * \left(\frac{\sum_{i\neq j} miss(i,j)}{number of samples in class i}\right)\ \ \ \forall j \in C
\label{eq:mi}
\end{equation}
The computation of mass values for other elements of the power set such as $m_{ij},m_{ijk}...$ is more complicated. The confusion matrix provides us with information regarding misclassification among two classes as well. Higher combination of classes we not considered henceforth because they needlessly increase the computation while not providing significant information. In other words while considering the predicted class of a patch we are giving consideration to one more class which has a high probability of confusion with chosen class. So mass values such as $m_{ijk},m_{ijkl},...$ are ignored. If we remember from equation \ref{eq:mi} the probability of each class was deducted by a certain amount to obtain corresponding mass values. If all these deductions are accumulated and redistributed among other members of the power set as their mass values then the requisite of a mass distribution is satisfied which is given by $\sum_{i \in 2^C}m_i = 1$. Let the accumulated deductions be defined as $D$.

\begin{equation}
D = \sum_{i \in C} p_i * \left(\frac{\sum_{i\neq j} miss(i,j)}{number of samples  in  class i}\right)
\end{equation}

The mass values of higher order members of the power set with a cardinality of 2 is given by
\begin{equation}
m_{jk} = \frac{miss(j,k)+miss(k,j)}{\sum_{p,q \in C, p \neq q}(miss(p,q)+miss(p,q))}*D  
\end{equation}
After computing the mass distribution for each of the three CNNs we combine them to find the final mass distribution using the Dempster-Shafer rule of combination of evidence as described in section 2.2.1 of ~\cite{dempster}

\section{Experimentations}
The first part of our experimentation trains and tabulates the performance of the three individual CNNs. Optimum size for the raw superpixel patch and 1st and 2nd neighbour images were chosen as $24 \times 24$, $32 \times 32$ , and $48 \times 48$ respectively based on validation performance. The architecture and the corresponding performance is given in table \ref{tab:indiv}. The second phase records the result of the various ensemble methods. The ICCV 09 Database was used for the experimentation. It contains 715 images with ground truths showing 8 semantically segmented classes. The dataset was split into 500 training, 72 validation and 143 test samples. The minimum object size considered for generating the superpixels was approximately $ 20 \times 20$. The total number of superpixels was 291,911.

\section{Results and analysis}
\begin{figure}[htbp]
\centering
\includegraphics[width=\textwidth]{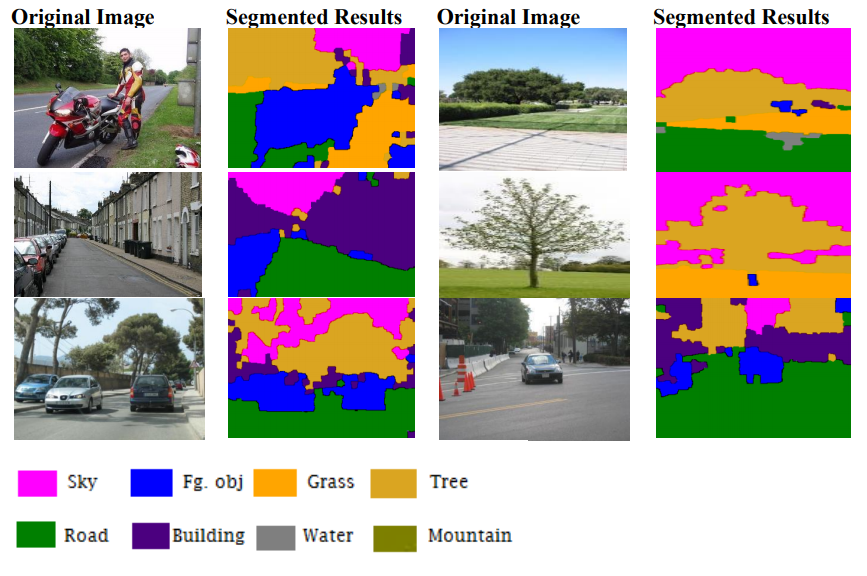}
\caption{Segmentation Results for our proposed approach}
\label{results}
\end{figure}

In figure \ref{results} we can see some segmented examples as generated by our approach. In the next sub-sections we shall look into the performance of the individual CNNs and how they improved upon using ensemble techniques.

\subsection{Individual CNNs}
Each individual CNN was trained over 500 images. The optimal architecture was selected according to their performance on the validation dataset. The final test accuracy along with the optimum configurations is shown in table \ref{tab:indiv}.
\begin{table}[t]
\centering
\caption{Classification performance of Individual CNNs. $nCk=n$ number of $k \times k$ convolution kernels, $mP=Max-pooling$ with $m \times m$ window and $stride = m$, $FCp=Fully\ connected\ layer$ with $p$ units, $qN$ refers to input patch along with $qth$-level neighbours}
\label{tab:indiv}
\begin{tabular}{|c|c|c|c|}
\hline
\textbf{Patch} & \textbf{\begin{tabular}[c]{@{}c@{}}Input \\ Size\end{tabular}} & \textbf{\begin{tabular}[c]{@{}c@{}}Network \\ Architecture\end{tabular}} & \textbf{\begin{tabular}[c]{@{}c@{}}Test Classification\\  Accuracy\end{tabular}} \\ \hline
0N & $24 \times 24$ & 32C5-2P-64C3-2P-FC256 & 72.32 \\
1N & $32 \times 32$ & 32C7-2P-64C5-2P-FC256 & 72.45 \\
2N & $48 \times 48$ & 32C7-2P-64C5-2P-FC256 & 72.24 \\ \hline
\end{tabular}
\end{table}

It can be seen that all the individual CNNs perform almost at the same level. Thus it may seem that the choice of different neighbours is ineffective. However if we ensemble the softmax outputs of these three CNNs we see a different story.

\subsection{Ensemble Methods}
We chose three ensemble strategies to deal with disagreement among the individual CNNs along with Dempster-Shafer theory to remove uncertainty in the obtained results. Table \ref{tab:ensemble} shows performance of the three ensemble strategies across all classes. 

\begin{table}[htbp]
\centering
\caption{Performance of Ensemble approaches with respect to various classes}
\label{tab:ensemble}
\begin{tabular}{|l|c|c|c|c|c|c|c|c|c|}
\hline
\multicolumn{1}{|c|}{\multirow{2}{*}{\textbf{Type}}} & \multicolumn{8}{c|}{\textbf{Test accuracy}} & \multirow{2}{*}{\textbf{\begin{tabular}[c]{@{}c@{}}Avg.\\ Acc.\end{tabular}}} \\ \cline{2-9}
\multicolumn{1}{|c|}{} & \multicolumn{1}{l|}{\textbf{Sky}} & \multicolumn{1}{l|}{\textbf{Tree}} & \multicolumn{1}{l|}{\textbf{Grass}} & \multicolumn{1}{l|}{\textbf{Ground}} & \multicolumn{1}{l|}{\textbf{Building}} & \multicolumn{1}{l|}{\textbf{Mountain}} & \multicolumn{1}{l|}{\textbf{Water}} & \multicolumn{1}{l|}{\textbf{Object}} &  \\ \hline
\textit{\begin{tabular}[c]{@{}l@{}}Dempster\\ Shafer\end{tabular}} & \textbf{88.07} & 76.74 & \textbf{80.29} & \textbf{86.59} & \textbf{77.8} & 2.56 & 59.41 & 59.18 & 77.07 \\ \hline
\textit{\begin{tabular}[c]{@{}l@{}}Max\\ Voting\end{tabular}} & 84.05 & 73.37 & 73.65 & 80.80 & 69.90 & 4.15 & 63.90 & 60.79 & 72.88 \\ \hline
\textit{\begin{tabular}[c]{@{}l@{}}Weighted\\ Average\end{tabular}} & 87.75 & \textbf{77.17} & 79.69 & 85.43 & 75.85 & \textbf{4.16} & \textbf{64.69} & \textbf{63.24} & \textbf{77.14} \\ \hline
\end{tabular}
\end{table}
It can bee seen that for some classes Dempster-Shafer wins, whereas for other classes the weighted average is ahead. The poor performance in the mountain category was due to the fact that segments with mountains were quite scarce throughout the dataset.

Finally, in table \ref{tab:comp} we can see how our approach performs against some fantastic works on the database. As it can be seen all the other works that has been compared with are from world class conferences and journals. Our approach was able to beat all of them.

\begin{table}[htbp]
\centering
\caption{Our approach compared with other approaches on the ICCV09 dataset}
\label{tab:comp}
\begin{tabular}{|c|c|c|}
\hline
\textbf{Approaches} & \textbf{Methodology} & \textbf{\begin{tabular}[c]{@{}c@{}}Classification\\ Accuracy\end{tabular}} \\ \hline
\textit{Baseline (ICCV 09)~\cite{ICCV09}} & Pixel CRF & 74.3 \\
\textit{Gould et al. (ICCV 09) ~\cite{ICCV09}} & Region based energy & 76.4 \\
\textit{Munoz et al. (ECCV 10) ~\cite{munoz}} & Probabilistic Model & 76.9 \\
\textit{Farabetet al. (PAMI 13) ~\cite{farabet}} & CNN+Superpixel & 74.56 \\
\textit{Tighe et al. (ECCV 10) ~\cite{tighe}} & Features+Superpixel & 76.3 \\ \hline
\textit{\textbf{Our Approach}} & \textbf{Superpixel + CNN + Ensemble} & \textbf{77.14} \\ \hline
\end{tabular}
\end{table}
Moreover, the testing time has been calculated to be in the range of $25-30 ms$ on a GTX 1080 GPU. Thus ensuring successful real-time implementation.

\section{Conclusion}
We have implemented a novel approach for superpixel-level segmentation and boosted its performance by various ensemble methods and uncertainty handling. Our approach shows a fast method for creating decent segments. When compared with other methods that were applied in this dataset it showed its strength. In the future it is possible to extend this work to video segmentations. Overall we believe that speed the algorithm combined with a relatively small sized CNN our approach shows promising results.

\section*{Acknowledgment}

This work is partially supported by the project entitled ``Development of knowledge graph from images using deep learning'' sponsored by SERB (Government of India, order no. SB/S3/EECE/054/2016) (dated 25/11/2016), and carried out at the Centre for Microprocessor Application for Training Education and Research, CSE Department, Jadavpur University.

\bibliographystyle{spmpsci}
\bibliography{references}

\end{document}